# Optimization of Structured Mean Field Objectives

**Alexandre Bouchard-Côté***        **Michael I. Jordan*,†**
*Computer Science Division        †Department of Statistics
University of California at Berkeley

## Abstract

In intractable, undirected graphical models, an intuitive way of creating structured mean field approximations is to select an acyclic tractable subgraph. We show that the hardness of computing the objective function and gradient of the mean field objective qualitatively depends on a simple graph property. If the tractable subgraph has this property—we call such subgraphs $v$-*acyclic*—a very fast block coordinate ascent algorithm is possible. If not, optimization is harder, but we show a new algorithm based on the construction of an auxiliary exponential family that can be used to make inference possible in this case as well. We discuss the advantages and disadvantages of each regime and compare the algorithms empirically.

## 1 Introduction

The mean field approach to inference (Peterson and Anderson 1987) is a venerable framework for approximating partition functions and marginal probabilities of collections of random variables. Compared to its more recent variational cousins—Bethe entropy approximations (Yedidia et al. 2001), planar graph decompositions (Globerson and Jaakkola 2006), log-determinant relaxation (Wainwright and Jordan 2006), conic programming approaches (Lasserre 2000), etc—the mean field approach still retains some advantages. It combines simplicity with an important bounding property (it provides a lower bound on the partition function) and for this reason it is still widely used in applications (Hua and Wu 2006, Blei and Jordan 2005). Moreover, while the mean field approach yields a globally consistent set of moments, alternatives such as Bethe approximations generally do not. One consequence is that for mean field there is a canonical way of extracting samples from mean field approximations, a useful property when considering combinations of variational and MCMC sampling algorithms (Freitas et al. 2001).

An important extension to the basic ("naive") mean field approach was introduced in Saul and Jordan (1996) under the name of "structured mean field," with further developments due to a variety of authors, including Barber and Wiegerinck (1999) and Wiegerinck (2000). The structured mean field approach permits a trade-off between quality and computational cost. In the framework of graphical models, the structured mean field approach amounts to taking a subgraph of the original graphical model as an approximating family, where the subgraph is chosen so that inference is tractable (it is generally an acyclic subgraph). The structured mean field variational procedure is then to find the closest distribution in this sub-graphical model to the original distribution, as measured in KL divergence. In particular, naive mean field can be seen as the special case in which the tractable subgraph is equal to the completely disconnected graph.

The structured mean field variational problem is generally a nonconvex optimization problem. No matter what algorithm is used to perform the optimization, a prerequisite is that evaluation of the objective function and ideally of the gradient of the mean field objective should be fast. In this work, we show that an important aspect of this problem has been overlooked: we show that not all tractable subgraphs are computationally equivalent for the purposes of structured mean field approximation. In particular, we show that there are two distinct classes of such subgraphs—that we refer to as $v$- and $b$-acyclic—with their own distinct computational properties. Most tractable subgraphs found in previous work on structured mean field approximation fall in the first category. In the second category, computation is more challenging—but there is a significant gain in accuracy. To reap the benefits of this gain in accuracy, we present a new structure mean field algorithm based on auxiliary exponential families.

The notion of $v$- and $b$-acyclic subgraphs is different that the notion of "overlapping cluster" used in the work on structured mean field by Geiger et al. (2006). Some $v$-acyclic graphs have overlapping clusters, some do not; and moreover, the computational dichotomy we establish here does not hold if the notion of $v$- and $b$-acyclic subgraphs is replaced by that of overlapping clusters. Note that other variational approximations such as Expectation Propagation also have a subgraph interpretation (Minka and Qi 2003). While this subgraph sometimes happens to be $b$-acyclic, there is no special distinction between $v$- and $b$-acyclic graphical approximations in the case of Bethe-energy variational approximations. This is why we focus on mean field approximations.

The paper is organized as follows. We present a basic introduction to structured mean field in Section 2. We then discuss our analysis and algorithmic developments in Section 3. We present empirical results to support our claims in Section 4 and we present our conclusions in Section 5.

## 2 Background

In this section, we review the principles of mean field approximation and set the notation. Our exposition follows the general treatment of variational methods presented in Wainwright and Jordan (2003) where the Legendre-Fenchel transformation plays a central role.

### 2.1 Exponential families

We assume that the random variable under study, $X_{\boldsymbol{\theta}}$, has a distribution in a regular exponential family $\mathscr{P}$ in canonical form:

$$\mathbb{P}(\boldsymbol{X_\theta} \in A) = \int_A \exp\{\langle \boldsymbol{\phi}(x), \boldsymbol{\theta}\rangle - A(\boldsymbol{\theta})\}\nu(\mathrm{d}x), \quad (1)$$

$$A(\boldsymbol{\theta}) = \log \int \exp\{\langle \boldsymbol{\phi}(x), \boldsymbol{\theta}\rangle\}\nu(\mathrm{d}x) \quad (2)$$

for a sufficient statistics $\boldsymbol{\phi}: \boldsymbol{\mathcal{X}} \to \mathbb{R}^d$, base measure $\nu$ and parameters $\boldsymbol{\theta} \in \Omega = \{\boldsymbol{\theta} \in \mathbb{R}^d : A(\boldsymbol{\theta}) < \infty\}$.

We will also use the notation $X_{\boldsymbol{\mu}}$ where $\boldsymbol{\mu} \in \mathbb{R}^d$ to denote a random variable with distribution in $\mathscr{P}$ such that $\mathbb{E}[\boldsymbol{\phi}(X_{\boldsymbol{\mu}})] = \boldsymbol{\mu}$. Note that this is well defined since $\boldsymbol{\phi}$ is sufficient for $\theta$.

We are interested in the case in which the distribution of $\boldsymbol{X}$ factors according to an undirected graphical model on $m$ vertices $G = (V, E)$, i.e. $\boldsymbol{X} = (X_1, \ldots, X_m)$, $\boldsymbol{\mathcal{X}} = \mathcal{X}^m$. For simplicity of notation we focus on the case in which the interactions are pairwise and the base measure is discrete. However, the ideas apply directly to the general exponential family—this will be discussed in more detail in Section 3.

Let $F = (V \times \mathcal{X}) \cup (E \times \mathcal{X}^2)$ be the index set for the coordinates of $\boldsymbol{\phi}$ (the *potentials*). If $e = (a, b) \in E$, then it is understood that the following inclusion holds on the induced sigma-algebra: $\sigma(\phi_{e,\cdot}(\boldsymbol{X})) \supseteq \sigma(X_a, X_b)$. Similarly, if $v \in V$, $\sigma(\phi_{v,\cdot}(\boldsymbol{X})) \supseteq \sigma(X_a)$. We lose no generality by requiring existence of potentials for all vertices and edges, since we can always set their corresponding parameter to zero.

### 2.2 Convex duality

A simple but fundamental property of exponential families is that the gradient and Hessian of the log partition function have the following forms:

$$\nabla A(\boldsymbol{\theta}) = \mathbb{E}[\boldsymbol{\phi}(\boldsymbol{X_\theta})]$$
$$H(A(\boldsymbol{\theta})) = \mathbf{V}ar[\boldsymbol{\phi}(\boldsymbol{X_\theta})]. \quad (3)$$

The second identity implies convexity, which we can use in conjunction with the Legendre-Fenchel transformation to establish an alternative form for $A$.

**Definition 1** *For an extended real-valued function $f$, the Legendre-Fenchel transformation is defined as:*

$$f^*(x) = \sup\{\langle x, y\rangle - f(y) : y \in \mathrm{dom}(f)\}.$$

When $f$ is convex and lower semi-continuous, $f = f^{**}$, we can use convexity of $A$ to obtain:

$$A(\boldsymbol{\theta}) = \sup\{\langle \boldsymbol{\theta}, \boldsymbol{\mu}\rangle - A^*(\boldsymbol{\mu}) : \boldsymbol{\mu} \in \mathscr{M}\}, \quad (4)$$

where $\mathscr{M} = \nabla A(\Theta)$ is the set of *realizable moments*.

Formulation (4) is no more tractable than the definition of $A$ in Equation (2): the term $A^*(\boldsymbol{\mu})$, which can be shown to be equal to the negative of the entropy $-H_\nu(\boldsymbol{X_\mu})$ (Wainwright and Jordan 2008) cannot be computed efficiently for arbitrary $\boldsymbol{\mu}$. Hence, the objective function cannot be evaluated efficiently. On the other hand, Equation (4) is a constrained optimization problem that can be relaxed. Mean field methods can be seen as a particular type of relaxation where the sup is taken over a proper subset of $\mathscr{M}$. In particular, the sup is taken over a subset of $\mathscr{M}$ for which the objective function and its gradient can be evaluated efficiently.

### 2.3 Graphical mean-field relaxations

In order to define this relaxation, the user needs to provide a subset of the principal exponential family,

$\mathcal{Q} \subset \mathcal{P}$, in which the log partition and moments can be computed.

An intuitively appealing approach to making this selection is to make use of the graphical representation of the exponential family and to choose a subset of the edges, $E' \subset E$, to represent a tractable subfamily. In particular, in defining this subfamily we retain only the potentials with indices

$$F' = \{f \in F : f = (v, \cdot) \text{ for } v \in V \text{ or } \\ f = (e, \cdot) \text{ for } e \in E'\}.$$

The subgraph $G' = (V, E')$ is generally taken to be acyclic so that inference in the induced subfamily $\mathcal{Q}$ is indeed tractable.

We denote the parameters indexing this subfamily by $\boldsymbol{\omega} \in \Xi$ and its moments by $\boldsymbol{\tau}$. Also we let $\boldsymbol{Y}$ denote a generic random variable that has a distribution in $\mathcal{Q}$. The subfamily induces a tractable subset $\mathcal{M}_{\text{MF}} \subseteq \mathcal{M}$ of moments in $\mathcal{M}$:

$$\mathcal{M}_{\text{MF}} = \Big\{\boldsymbol{\mu} \in \mathcal{M} : \exists \boldsymbol{\omega} \in \Xi \text{ s.t. } \mathbb{E}[\boldsymbol{\phi}(\boldsymbol{Y_\omega})] = \boldsymbol{\mu}\Big\},$$

which in turn induces a tractable relaxation:

$$\hat{A}(\boldsymbol{\theta}) = \sup\{\langle \boldsymbol{\theta}, \boldsymbol{\mu}\rangle - A^*(\boldsymbol{\mu}) : \boldsymbol{\mu} \in \mathcal{M}_{\text{MF}}\}. \quad (5)$$

Indeed, for all $\mu \in \mathcal{M}_{\text{MF}}$, $A^*(\boldsymbol{\mu})$ amounts to computing the entropy of a forest-shaped graphical model:

$$H_\nu(\boldsymbol{Y}) = \sum_{i \in \text{cc}(G')} \Big\{H_\nu(Y_i) + \sum_{j:j\neq i \sim j} H_\nu(Y_j|Y_{\text{pa}(j)})\Big\}.$$

Note that $\hat{A}(\boldsymbol{\theta}) \leq A(\boldsymbol{\theta})$, which is a very useful property when mean-field inference is used in the inner loop of EM (Wainwright and Jordan 2008). Moreover, if $E'' \subseteq E'$, with associated mean field approximation $\breve{A}(\boldsymbol{\theta})$, then $\breve{A}(\boldsymbol{\theta}) \leq \hat{A}(\boldsymbol{\theta}) \leq A(\boldsymbol{\theta})$.

As a consequence, adding edges in $G' = (V, E')$ can only increase the quality of the global optimum. This does not imply that the local optimum found by the optimization procedure will always be superior, but we show in the experimental section that empirically there is indeed an improvement when edges are added to the approximation.

We also let $\mathcal{N}$ denote the set of realizable moments of $\mathcal{Q}$. Note that this set is formally distinct from $\mathcal{M}_{\text{MF}}$ (in particular, its elements have different dimensionality).

By construction, it will be possible to perform the optimization over variables in $\mathbb{R}^{d'}$, where $d' = |F'|$. To see why, let us define an embedding $\boldsymbol{\Gamma} : \mathbb{R}^{d'} \to \mathbb{R}^{\Delta d}$ ($\Delta d = d - d'$) as follows: for $f \in F \backslash F'$,

$$\Gamma_f(\boldsymbol{\tau}) = \mathbb{E}[\boldsymbol{\phi}_f(\boldsymbol{Y_\tau})].$$

We can then write the fundamental equation of mean-field approximation:

$$\sup\{\langle \boldsymbol{\theta}, \boldsymbol{\mu}\rangle - A^*(\boldsymbol{\mu}) : \boldsymbol{\mu} \in \mathcal{M}_{\text{MF}}\} = \\ \sup\{\langle \boldsymbol{\omega}, \boldsymbol{\tau}\rangle + \langle \boldsymbol{\vartheta}, \boldsymbol{\Gamma}(\boldsymbol{\tau})\rangle - A^*(\boldsymbol{\tau}) : \boldsymbol{\tau} \in \mathcal{N}\}, \quad (6)$$

where $\boldsymbol{\theta} = (\boldsymbol{\omega}, \boldsymbol{\vartheta})$. Note the slight abuse of notation: we use $A$ to denote the partition function of both exponential families; the notation can always be disambiguated by inspecting the dimensionality of the parameter vector.

The right-hand side of Equation (6) makes it clear that the mean field optimization problem is different than performing inference in $\mathcal{Q}$, the latter being:

$$\sup\{\langle \boldsymbol{\omega}, \boldsymbol{\tau}\rangle - A^*(\boldsymbol{\tau}) : \boldsymbol{\tau} \in \mathcal{N}\}.$$

In particular, the function $\boldsymbol{\Gamma}$ on the right-hand side of Equation (6) is generally non-convex. The precise form of $\boldsymbol{\Gamma}$ will be established shortly.

The left-hand side of Equation (6) gives another perspective on the mean-field optimization problem: here we have a convex objective, but the optimization is over a non-convex set (Wainwright and Jordan 2008).

Note that Equation (6) allows us to perform the optimization in the smaller space $\mathbb{R}^{d'}$; this is a key algorithmic consequence of the mean-field approximation.

### 2.4 Generic fixed point updates

Let $G(\boldsymbol{\tau}) = \langle \boldsymbol{\omega}, \boldsymbol{\tau}\rangle + \langle \boldsymbol{\vartheta}, \boldsymbol{\Gamma}(\boldsymbol{\tau})\rangle - A^*(\boldsymbol{\tau})$. We take partial derivatives to obtain stationary point conditions. By the definition of $\boldsymbol{\Gamma}$:

$$\frac{\partial G}{\partial \tau_f}(\boldsymbol{\tau}) = \omega_f + \sum_{g \in F \backslash F'} \vartheta_g \frac{\partial \Gamma_g}{\partial \tau_f}(\boldsymbol{\tau}) - \frac{\partial A^*}{\partial \tau_i}(\boldsymbol{\tau})$$

where $f \in F'$.

It will be useful to represent this update in vector notation; for this purpose, we introduce the following definition.

**Definition 2** *The* embedding Jacobian *is the (transposed) Jacobian matrix of* $\boldsymbol{\Gamma}$:

$$J = \Big(\frac{\partial \Gamma_g}{\partial \tau_f}\Big)_{f,g}$$

for $f \in F', g \in F \backslash F'$.

With this definition, we obtain the concise expression:

$$\nabla G = \boldsymbol{\omega} + J\boldsymbol{\vartheta} - \nabla A^*.$$

A necessary condition for optimality is therefore:

$$\nabla A^* = \boldsymbol{\omega} + J\boldsymbol{\vartheta} \quad (7)$$

Since $\nabla A^*$ is injective (Wainwright and Jordan 2008), $\nabla A(\nabla A^*(\tau)) = \tau$, we obtain the fixed point equation:

$$\tau = \nabla A\Big(\omega + J(\tau)\vartheta\Big),$$

and it is this equation that must be solved in the mean-field approximation.

Now that we have a simple formula for the mean field updates, we will look at how difficult the computation of the entries in $J$ are.

## 3 The embedding Jacobian

The hardness of the mean-field optimization is determined by the nature of the embedding Jacobian matrix $J(\tau)$. We now show that the form of this matrix is governed by a simple graph property.

**Definition 3** *An acyclic subgraph with edges $E' \subseteq E$ is called v-acyclic (very acyclic) if, for all $e \in E$, $E' \cup \{e\}$ is still acyclic. Otherwise it is called b-acyclic (barely acyclic).*

This definition can be generalized easily to the case of factor graphs. In this case, we ignore degenerate cycles involving a number of variables smaller than the largest connectivity of the factor visited, but require that for all pairs $e, e' \in E$, that $E' \cup \{e, e'\}$ be acyclic.

### 3.1 $v$- and $b$-acyclic subgraphs

We now present some examples of $v$- and $b$-acyclic subgraphs to illustrate the definition.

We start with the important example of $n$-dimensional Ising models, where vertices are arranged in a regular lattice. Naive mean field corresponds to taking the completely disconnected graphical model. A more structured approximation can be obtained by using a spanning forest. However, not all such subgraphs are $v$-acyclic. We show in Figure 1 (left column) examples of subgraphs that do and do not have this property.

Next, to illustrate the definition in the factor graph case, consider a factorial HMM that consists of $k$ chain graphs of the same lengths, with all vertices at position $i$ in each chain linked together by a factor. This is equivalent to the factor graph with one pairwise factor between contiguous vertices and a $k$-wise factor at each position (see Figure 1, right column). (This coupling is typically induced by a v-structure in an underlying directed graphical model.) Previous work on structured mean field has used the approximation formed by $k$ independent chains. This is $v$-acyclic, since adding any two edges in the graphs does not create non-degenerate cycles. Note however that a slightly more structured approximation—containing a single extra factor as shown in Figure 1 (bottom right)—is not. This is an example where a small difference in the structured mean field subgraph has a big impact on the complexity of inference.

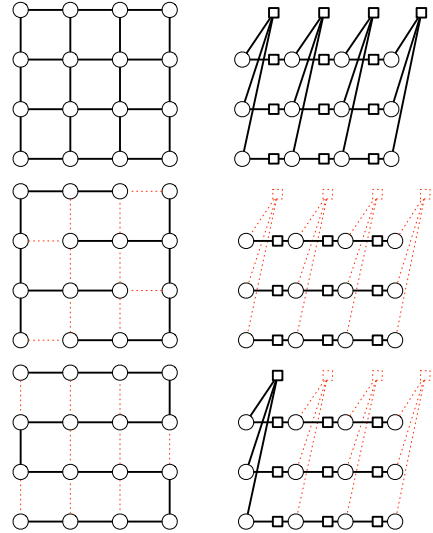

Figure 1: Examples of $v$-acyclic and $b$-acyclic subgraph. *Left column, from top to bottom:* the original graphical model $G$ (a two-dimensional Ising model); a $v$-acyclic tractable subgraph $G'$ (dashed lines are those in $E \backslash E'$); a $b$-acyclic subgraph. *Right column, from top to bottom:* the factor graph representation of a factorial HMM; a $v$-acyclic factor subgraph; a $b$-acyclic factor subgraph.

### 3.2 Connected components decomposition of the embedding Jacobian

In order to explain the consequence of the $v$-acyclic vs. $b$-acyclic distinction on the form of the embedding Jacobian, it will be useful to first decompose $G'$ into connected components $G'^{(1)}, \ldots, G'^{(k)}$, where each $G'^{(c)} = (V^{(c)}, E'^{(c)})$ is a connected subgraph such that $\cup_c G'^{(c)} = G'$. Each connected component $G'^{(c)}$ induces a subset of the coordinates $F'^{(c)}$, and induces an exponential family $\mathscr{Q}^{(c)}$ with corresponding parameters $\omega^{(c)}$ where $\omega = (\omega^{(1)}, \ldots, \omega^{(k)})$ and partition function $A^{(c)}$.

**Proposition 4** *We have:*

$$\tau = \begin{pmatrix} \tau^{(1)} \\ \vdots \\ \tau^{(k)} \end{pmatrix} = \begin{pmatrix} \nabla A^{(1)}\big(\omega^{(1)} + J^{(1)}(\tau)\vartheta\big) \\ \vdots \\ \nabla A^{(k)}\big(\omega^{(k)} + J^{(k)}(\tau)\vartheta\big) \end{pmatrix}$$

*where $J^{(c)}$ is the restriction of $J$ to partial derivatives indexed in the corresponding connected component.*

This proposition is proved by writing the log partition function of $\mathcal{Q}$ in terms of the following decomposition:

$$A(\zeta) = \log \int \exp\{\langle \zeta^{(1)}, \phi^{(1)}(x) \rangle + \ldots$$
$$+ \langle \zeta^{(k)}, \phi^{(k)}(x) \rangle\}\nu(\mathrm{d}x)$$
$$= \sum_{c \in \mathrm{cc}(G')} \log \int \exp\langle \zeta^{(c)}, \phi^{(c)}(x) \rangle \nu(\mathrm{d}x)$$
$$= \sum_{c \in \mathrm{cc}(G')} A^{(c)}(\zeta^{(c)}),$$

valid for any $\zeta \in \Xi$. We therefore have:

$$\nabla A(\zeta) = \begin{pmatrix} \nabla A_1(\zeta_1) \\ \vdots \\ \nabla A_k(\zeta_k). \end{pmatrix}$$

Now plug in $\zeta^{(c)} = \omega^{(c)} + J^{(c)}(\tau)\vartheta$.

### 3.3 Optimization of $v$-acyclic components

Suppose now that connected component $G'^{(c)}$ is $v$-acyclic. We show how entries in the embedding Jacobian $J^{(c)}(\tau)$ can be computed in constant time in this case.

Recall that we have:

$$J^{(c)}_{f,g}(\tau) = \frac{\partial}{\partial \tau_f} \mathbb{E}[\phi_g(\boldsymbol{Y}_\tau)],$$

where $f \in F'^{(c)}, g \in F \setminus F'$.

Since $g \in F \setminus F'$, we must have $g = (e, (s,t))$ for some $e = (a,b) \in E \setminus E'$ and $s,t \in \mathcal{X}^2$. Therefore:

$$J^{(c)}_{f,g}(\tau) = \frac{\partial}{\partial \tau_f} \mathbb{P}(Y_a = s, Y_b = t),$$

where $\boldsymbol{Y}_\mu = (Y_1, \ldots, Y_m)$.

There are three subcases to consider:

1. *Exactly one of the vertices $a,b$ belongs to $V^{(c)}$:* Suppose $a \in V^{(c)}, b \notin V^{(c)}$. Since $a,b$ are in different connected components, then by the Hammersley-Clifford theorem they are independent, so that:

$$J^{(c)}_{f,g}(\tau) = \frac{\partial}{\partial \tau_f} \mathbb{P}(Y_a = s)\mathbb{P}(Y_b = t)$$
$$= \frac{\partial}{\partial \tau_f} \tau_{a,s} \tau_{b,t}$$
$$= \tau_{b,t} \times \mathbf{1}[f = (a,s)]$$

2. *Both $a$ and $b$ belong to $V^{(c)}$:* We claim that this cannot, in fact, occur in $v$-acyclic components. Suppose the contrary: since $a,b$ belong to the same connected component, there is a path between $a$ and $b$. This means that $E'^{(c)} \cup \{(a,b)\}$ has a cycle, a contradiction.

3. *Neither $a$ nor $b$ belong to $V^{(c)}$:* In this case $J_{f,g}(\tau) = 0$.

Hence the number of nonzero entries in $J^{(c)}$ is equal to the number of edges in $E \setminus E'$ that have an endpoint in $V^{(c)}$ and each entry is a quantity that does not depend on $\tau^{(c)}$. This shows that:

**Proposition 5** *The right hand side of the equation*

$$\tau^{(c)} = \nabla A^{(c)}\big(\omega^{(c)} + J^{(c)}(\tau^{(1)}, \cdots, \tau^{(k)})\vartheta\big)$$

*does not depend on $\tau^{(c)}$. It is hence possible to optimize exactly in time $O(|F'^{(c)}|)$ the block of coordinates $\tau^{(c)}$ while keeping the values of the other blocks fixed.*[1]

An interesting property of this coordinate ascent algorithm is that it is guaranteed to converge to a local optimum (Wainwright and Jordan 2008). This can be seen from Equation (6) by using the fact that the original problem in Equation (4) is convex.

### 3.4 Relation to Gibbs sampling

Before moving to the $b$-acyclic case, we draw a parallel between block Gibbs sampling and structured mean field approximations in the case in which all connected components $G'^{(c)}$ are $v$-acyclic. This connection generalizes the classical relationship between naive Gibbs sampling steps and naive mean field coordinate ascent updates.

Recall that for binary random variables $X_s$ with coupling weights $\theta_{s,s'}$ and observation weights $\theta_s$, the naive Gibbs sampler is defined as the Markov chain $\boldsymbol{X}(1), \boldsymbol{X}(2), \ldots$ where only coordinate $s_t$ is resampled at time $t$ with transition probabilities:

$$X_{s_t}(t) \Big| \boldsymbol{X}(t-1) \sim$$
$$\mathrm{Bernoulli}\Big(\sigma\big[\theta_{s_t} + \sum_{s' \in N(s_t)} \theta_{s',s_t} X_{s'}(t-1)\big]\Big),$$

where $\sigma(x) = \{1 + \exp(-x)\}^{-1}$. This closely resembles the naive mean field coordinate updates:

$$\mu_{s_t}(t) \leftarrow \sigma\big[\theta_{s_t} + \sum_{s' \in N(s_t)} \theta_{s',s_t} \mu_{s'}(t-1)\big].$$

Now the parallel between block Gibbs sampling and $v$-acyclic structured mean fields is the following: let $\mathrm{MRF}(\mathcal{P}, \boldsymbol{\theta})$ denote the distribution of $\boldsymbol{X}_{\boldsymbol{\theta}}$. Then one can check easily that the block Gibbs sampler with blocks $V^{(1)}, \ldots, V^{(k)}$ has a transition kernel given by:

$$\boldsymbol{X}^{(c_t)}(t)\Big|\boldsymbol{X}(t-1) \sim$$
$$\mathrm{MRF}\left(\mathscr{P}'^{(c_t)}, \boldsymbol{\omega}^{(c_t)} + B^{(c_t)}(\boldsymbol{X}(t-1))\boldsymbol{\vartheta}\right),$$

where $B$ is the matrix defined as follows:

$$B_{f,g}^{(c)}(\boldsymbol{X}(t-1)) = \mathbf{1}[f = (a,s)] \times$$
$$\mathbf{1}[|\{a,b\} \cap V^{(c)}| = 1] \times$$
$$\mathbf{1}[X_b(t-1) = s],$$

and where $f \in F'^{(c)}, g \in F \backslash F'$ and $((a,b),(s,t)) = g$.

---

[1] Since we are treating graphical models that have only pairwise potentials, we drop from now on the quadratic dependency on the random variable state spaces, $|\mathcal{X}|^2$.

Note that the sparsity pattern of $B$ follows that of $J$. Moreover, the complexity of sampling from $\mathrm{MRF}(\mathscr{P}'^{(c)}, \cdot)$ is the same (up to a multiplicative constant) as the complexity of computing $\nabla A^{(c)}(\cdot)$. Both require executing sum-product on the same tree.

This parallel breaks when one or several of the connected components is $b$-acyclic. In this case there is no corresponding tractable block Gibbs sampler, while (as we discuss in the following section) mean field is still tractable albeit at a computational cost higher than in the $v$-acyclic case.

### 3.5 Optimization of $b$-acyclic components

We now turn to the case of structured mean field when there are $b$-acyclic components.

To simplify the notation we assume that there is a single $b$-acyclic component that spans the entire graph $G$. The derivation is essentially identical when there are several connected components.

The quantity to compute is:

$$J_{f,g}(\boldsymbol{\tau}) = \frac{\partial}{\partial \tau_f} \mathbb{P}(Y_a = s, Y_b = t),$$

where this time the probability on the right-hand side cannot be decoupled into a product of marginals. Let

$$\boldsymbol{p}_g = \{a = p_0, p_1, \ldots, p_k = b : \forall i, (p_i, p_{i+1}) \in E'\}$$

denote the shortest path in $G'$ from $a$ to $b$ (see Figure 2).

If we let $y_0 = s, y_k = t$, we have:

$$J_{f,g}(\boldsymbol{\tau}) = \frac{\partial}{\partial \tau_f} \sum_{y_1 \in \mathcal{X}} \cdots \sum_{y_{k-1} \in \mathcal{X}} \mathbb{P}(Y_{p_i} = y_i, \forall i \in \{1, \ldots k\})$$
$$= \frac{\partial}{\partial \tau_f} \mathbb{P}(Y_a = s) \sum_{y_1 \in \mathcal{X}} \mathbb{P}(Y_{p_1} = y_1 | Y_{p_0} = y_0) \sum_{y_2 \in \mathcal{X}} \cdots$$
$$= \frac{\partial}{\partial \tau_f} \tau_{a,s} \sum_{y_1 \in \mathcal{X}} \frac{\tau_{(p_0,p_1),(y_0,y_1)}}{\tau_{p_0,y_0}} \sum_{y_2 \in \mathcal{X}} \cdots$$

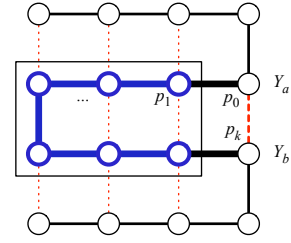

Figure 2: An example of the notation used in this section. This corresponds to the inference problem in the mean field subgraph in Figure 1, left column, bottom row. The box indicate which nodes are involved in the auxiliary exponential family $\mathscr{P}^{[g]}$ used to compute $J_{f,g}$ for all $f$. The edges in the path $\boldsymbol{p}_g$ are in bold and the edge corresponding to $g$ is the bold dashed line between $Y_a$ and $Y_b$.

From this expansion, we see immediately that $J_{f,g}(\boldsymbol{\tau})$ will not have the same sparsity properties as in the $v$-acyclic case.

One way to find these partial derivatives is to construct a specialized dynamic programming algorithm. This task is non-trivial: to see why, notice that when the partial derivative is taken with respect to a coordinate $\tau_f$ corresponding to an edge in the path $\boldsymbol{p}_g$, there are factors $\tau_f$ that appear both in the numerator and denominator:

$$J_{f,g}(\boldsymbol{\tau}) = \frac{\partial}{\partial \tau_f}\Bigg\{\bigg(\sum_{s'} \tau_{(a,p_1),(s,s')}\bigg) \sum_{y_1 \in \mathcal{X}} \frac{\tau_{(p_0,p_1),(y_0,y_1)}}{\left(\sum_{s'} \tau_{(p_0,p_1),(y_0,s')}\right)}$$
$$\times \sum_{y_2 \in \mathcal{X}} \cdots \sum_{y_{k-1} \in \mathcal{X}} \frac{\tau_{(p_{k-2},p_{k-1}),(y_{k-2},y_{k-1})}}{\left(\sum_{s'} \tau_{(p_{k-2},p_{k-1}),(y_{k-2},s')}\right)}$$
$$\times \frac{\tau_{(p_{k-1},p_k),(y_{k-1},y_k)}}{\left(\sum_{s'} \tau_{(p_{k-1},p_k),(y_{k-1},s')}\right)}\Bigg\}$$

The task is thus more complex than that of creating a dynamic programming algorithm of the type used for sum-product. A chain rule would need to be used, changing the form of the recursion for each $f, g$. The complexity of this naive approach can be shown to be $O(|E'| \times |F'| \times |F \backslash F'|)$, which is considerable.

We now present an alternative approach that is both simpler to implement and asymptotically faster. The idea is to construct an auxiliary exponential family model and to use an elementary property of Jacobian matrices to reduce the computation to a standard application of sum-product in the auxiliary exponential family.

There is one auxiliary exponential family $\mathscr{P}^{[g]}$ for each $((a,b),(s,t)) = g \in F \backslash F'$. It is defined on the chain $p_1, p_2, \ldots, p_{k-1}$ where $\boldsymbol{p}_g$ is as above. We pick its parameters $\boldsymbol{\theta}^{[g]}$ so that the partial derivatives of its par-tition function coincide with the quantity of interest:

$$Z^{[g]}(\boldsymbol{\theta}^{[g]}) = \sum_{\boldsymbol{x} \in \mathcal{X}^{k-1}} \exp\{\langle \boldsymbol{\phi}(\boldsymbol{x}), \boldsymbol{\theta}^{[g]} \rangle\}$$

$$= \Big(\sum_{s'} \tau_{(a,p_1),(s,s')}\Big) \sum_{y_1 \in \mathcal{X}} \frac{\tau_{(p_0,p_1),(y_0,y_1)}}{\big(\sum_{s'} \tau_{(p_0,p_1),(y_0,s')}\big)}$$

$$\times \sum_{y_2 \in \mathcal{X}} \cdots \sum_{y_{k-1} \in \mathcal{X}} \frac{\tau_{(p_{k-2},p_{k-1}),(y_{k-2},y_{k-1})}}{\big(\sum_{s'} \tau_{(p_{k-2},p_{k-1}),(y_{k-2},s')}\big)}$$

$$\times \frac{\tau_{(p_{k-1},p_k),(y_{k-1},y_k)}}{\big(\sum_{s'} \tau_{(p_{k-1},p_k),(y_{k-1},s')}\big)}$$

One can check that this is achieved with the following choice:

$$\boldsymbol{\theta}_h^{[g]} = \begin{cases} \log \tau_h - \log \tau_{v,x} + \log \tau_{(a,v),(s,x)} & \text{if } v = p_0 \\ \log \tau_h - \log \tau_{v,x} \\ \quad + \log \tau_{(p_{k-1},b),(y,t)} - \log \tau_{w,y} & \text{if } w = p_{k-1} \\ \log \tau_h - \log \tau_{v,x} & \text{otherwise} \end{cases}$$

where $h = ((v,w),(x,y))$, $(v,w) \in \boldsymbol{p}_g$, $(x,y) \in \mathcal{X}^2$.

Using Equation (3), we have for all $h = ((v,w),(x,y))$, $(v,w) \in \boldsymbol{p}_g, (x,y) \in \mathcal{X}^2$:

$$\frac{\partial Z^{[g]}}{\partial \theta_h^{[g]}} = Z^{[g]} \times \frac{\partial A^{[g]}}{\partial \theta_h^{[g]}} = Z^{[g]} \times \mu_h^{[g]},$$

where $A^{[g]} = \log Z^{[g]}$. This shows that one execution of sum-product at the cost of $O(|\boldsymbol{p}_g|)$ yields $|F'|$ entries.

Next, we define the following two Jacobian matrices:

$$I = \Big(\frac{\partial Z^{[g]}}{\partial \theta_h^{[g]}}\Big)_{g,h}; \quad K = \Big(\frac{\partial \theta_h^{[g]}}{\partial \tau_f}\Big)_{h,f}$$

where $I$ has size $|F \backslash F'| \times N$ and $K$ has size $N \times |F'|$, $N = \sum_{g \in F \backslash F'} |\boldsymbol{p}_g|$.

We can finally derive an expression for $J$, using the generalization of the chain rule for Jacobian matrices: $J = K^T I^T$.

After carrying this matrix multiplication, we obtain:

**Proposition 6** *When $G'$ is b-acyclic, the embedding Jacobian has the form:*

$$J_{f,g}(\boldsymbol{\tau}) = Z^{[g]} \times \begin{cases} \frac{\mu_{w,y}^{[g]}}{\tau_f} \times \mathbf{1}[x=s] & \text{if } v = p_0 \\ \mu_{v,x}^{[g]} \times \big\{ \frac{\mathbf{1}[y=t]}{\tau_f} - \frac{1}{\tau_{v,x}} \big\} & \text{if } w = p_{k-1} \\ \frac{\mu_f^{[g]}}{\tau_f} - \frac{\mu_{v,x}^{[g]}}{\tau_{v,x}} & \text{otherwise} \end{cases}$$

*where $f = ((v,w),(x,y))$ is such that $(v,w) \in \boldsymbol{p}_g$; otherwise, $J_{f,g}$ is equal to zero.*

The total cost of computing $J$ is $O(|E'| \times |F \backslash F'|)$, which is larger than the cost derived in the v-acyclic case, but smaller than the naive dynamic programming algorithm mentioned at the beginning of this subsection.

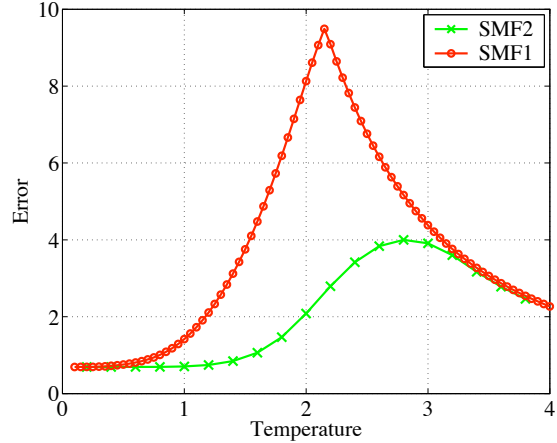

Figure 3: Error in the partition function estimate as a function of the temperature of the model. **SMF2** corresponds to a mean field approximation with more edges in $G'$ than SMF1: SMF1 is $v$-acyclic while SMF2 is not. See text for more details.

## 4 Experiments

We performed experiments on the $9 \times 9$ Ising model:

$$\phi((x_{i,j})_{i,j \in \{1,\ldots,9\}^2}) = \sum_{j=1}^{9} \sum_{i=1}^{8} \Big\{ x_{i,j} x_{i+1,j} + x_{j,i} x_{j,i+1} \Big\},$$

with $\theta_\star = \frac{1}{T}$, where $T$ is the *temperature* parameter as used in statistical physics. In this model, the partition function and moments can be computed exactly so that absolute errors can be established.

We used the updates of Proposition 5 when the tractable subgraph was $v$-acyclic and used the fixed point updates of Proposition 6 in the $b$-acyclic cases. Before performing the experiments, we verified empirically using directional derivatives that the updates that we have derived are error-free.

Since the computations in the $b$-acyclic case are more expensive, we first verified that they result in substantial gains in accuracy. For this purpose, we compared the following structured mean field approximations:

**SMF1** the approximation based on a $v$-acyclic subgraph with the smallest number of connected components possible (Figure 1, middle, left),

**SMF2** the approximation based on the graph shown in Figure 1, left column, bottom row—a $b$-acyclic subgraph.

In Figure 3 we show the results of this comparison at different temperatures. We see that there is indeed a regime, in this case around the phase transition point,

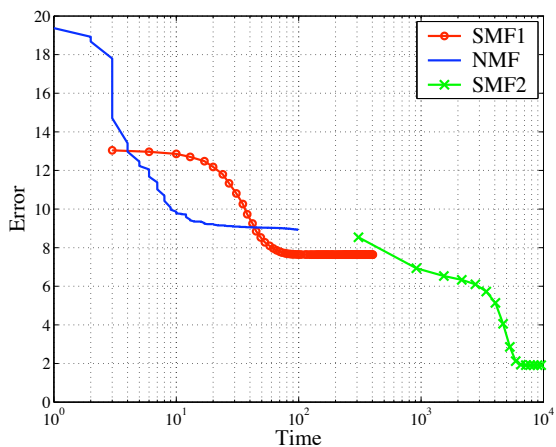

Figure 4: Error in the partition function estimate as a function of the running time in milliseconds (abscissa in a log scale) for three algorithms: naive mean field (NMF), $v$-acyclic structured mean field (SMF1) and $b$-acyclic mean field (SMF2).

where using the more expensive $b$-acyclic approximation pays off.

We also performed timing experiments to compare the convergence behavior of the mean field approximations SMF1 and SMF2. As a baseline, we ran the naive mean field (curve NMF on the graph). The results are displayed in Figure 4. We can see that in this model it takes one order of magnitude more time to move from naive mean field to $v$-acyclic structured mean field, and two orders of magnitude more time to move from $v$-acyclic to $b$-acyclic structured mean field approximations. This is consistent with the theoretical results developed in Section 3. Moreover, the bound on the log partition function gets tighter as more edges are added to the tractable subgraph.

## 5 Conclusion

We have characterized a dichotomy in the complexity of optimizing structure mean field approximations of graphical, exponential family models. The first class allows efficient block updates while the second is computationally more challenging. While most tractable subgraphs studied in the existing literature have fallen in the first category, we have presented theoretical and empirical reasons to expand the scope of the structured mean field method to consider the second category. We also presented a novel algorithm for computing the gradient and bound on the log-partition function in the $b$-acyclic case.